\pdfoutput=1

\documentclass[11pt]{article}

\usepackage{acl}

\newcommand{\sentscore}{\textbf{SenTScore}~}

\usepackage{times}
\usepackage{latexsym}
\usepackage{times}
\usepackage{latexsym}
\usepackage{amsmath}
\usepackage{multirow}
\usepackage{bbm}
\usepackage{graphicx}
\usepackage{amssymb}
\usepackage{booktabs}
\usepackage{adjustbox}
\usepackage{xspace}
\usepackage{algpseudocode}
\usepackage{algorithm}
\usepackage{graphicx}
\usepackage{caption}
\usepackage{subcaption}
\usepackage{color, colortbl}
\usepackage{tablefootnote}
\usepackage[explicit]{titlesec}
\usepackage{todonotes}
\usepackage{enumitem}
\usepackage{tablefootnote}
\usepackage[T1]{fontenc}
\usepackage[utf8]{inputenc}
\usepackage{microtype}

\title{Distillation of encoder-decoder transformers for sequence labelling}

\author{%
Marco Farina\thanks{\enspace Equal contribution} $^{1}$
~\;~\;~Duccio Pappadopulo\footnotemark[1] $^{1}$
~\;~\;~Anant Gupta$^{1}$ \\
\textbf{Leslie Huang}$^{1}$
~\;~\;~\textbf{Ozan İrsoy}$^{1}$~\;~\;~\textbf{Thamar Solorio}\thanks{\enspace Research completed during sabbatical at Bloomberg} $^{1,2}$\\
    $^{1}$Bloomberg~\;~\;~\;~$^{2}$Department of Computer Science, University of Houston\\
   \texttt{\small\{mfarina19,~dpappadopulo,~agupta968,~lhuang328,~oirsoy\}@bloomberg.net}\\
   \texttt{\small thamar.solorio@gmail.com}
 }

\begin{document}
\maketitle
\begin{abstract}
Driven by encouraging results on a wide range of tasks, the field of NLP is experiencing an accelerated race to develop bigger language models. 
This race for bigger models has also underscored the need to continue the pursuit of practical distillation approaches that can leverage the knowledge acquired by these big models in a compute-efficient manner. Having this goal in mind, we build on recent work to propose a hallucination-free framework for sequence tagging that is especially suited for distillation. We show empirical results of new state-of-the-art performance across multiple sequence labelling datasets and validate the usefulness of this framework for distilling a large model in a few-shot learning scenario.
\end{abstract}
\section{Introduction}

Sequence labelling (SL) can be defined as the task of assigning a label to a span in the input text. Some examples of SL tasks are: i) named entity recognition (NER), where these labelled spans refer to people, places, or organizations, and ii) slot-filling, where these spans or slots of interest refer to attributes relevant to complete a user command, such as \textit{song name} and \textit{playlist} in a dialogue system. In general, these spans vary semantically depending on the domain of the task.

Despite the strong trend in NLP to explore the use of large language models (LLMs) there is still limited work evaluating prompting and decoding mechanisms for SL tasks. In this paper we propose and evaluate a new inference approach for SL that addresses two practical constraints:
\begin{itemize}
    \item \textbf{Data scarcity:} The lack of vast amounts of annotated, and sometimes even the lack of unlabelled data, in the domain/language of interest. 
    \item \textbf{Restricted computing resources at inference time:} LLMs are very effective, but deploying them to production-level environments is expensive, especially in contexts with latency constraints, such as in a live dialogue system.
\end{itemize}

\begin{table*}[t]
\centering

  \setlength{\tabcolsep}{2pt}
  \resizebox{\textwidth}{!}{
  \begin{tabular}{llllll}
    \toprule
     \multicolumn{6}{l}{
     \textbf{Original text}
     } \\ \midrule

 play & wow & by & jon & theodore \\
 
    \toprule
    \multicolumn{6}{l}{ 
     \textbf{Encoder input for SenT$'$ format}
     }\\ \midrule
     
 \colorbox{lightgray}{{\fontfamily{cmtt}\selectfont<extra\_id\_0>}} play & \colorbox{lightgray}{{\fontfamily{cmtt}\selectfont<extra\_id\_1>}} wow & \colorbox{lightgray}{{\fontfamily{cmtt}\selectfont<extra\_id\_2>}} by & \colorbox{lightgray}{{\fontfamily{cmtt}\selectfont<extra\_id\_3>}} jon & \colorbox{lightgray}{{\fontfamily{cmtt}\selectfont<extra\_id\_4>}} theodore
 & \colorbox{lightgray}{{\fontfamily{cmtt}\selectfont<extra\_id\_5>}}\\

  	\toprule
  	    \multicolumn{6}{l}{ 
     \textbf{Expected decoder output for SenT$'$ format} 
     }\\ \midrule
  	 
  	 \colorbox{lightgray}{{\fontfamily{cmtt}\selectfont<extra\_id\_0>}} \color{cyan}O & \colorbox{lightgray}{{\fontfamily{cmtt}\selectfont<extra\_id\_1>}} \color{orange}TRACK &  \colorbox{lightgray}{{\fontfamily{cmtt}\selectfont<extra\_id\_2>}} \color{cyan}O & \colorbox{lightgray}{{\fontfamily{cmtt}\selectfont<extra\_id\_3>}} \color{orange}ARTIST & \colorbox{lightgray}{{\fontfamily{cmtt}\selectfont<extra\_id\_4>}} \color{orange}I &
         \colorbox{lightgray}{{\fontfamily{cmtt}\selectfont<extra\_id\_5>}} \\

    \bottomrule
  \end{tabular}
  }

        


        
 \caption{An example of how an original input text (from the SNIPS dataset) is transformed into the SenT$'$ input for the model, and the format for the expected output. We use the explicit form of the special token strings used by T5. The addition of the extra token at the end of the input differentiates SenT$'$ from SentT. Notice the modified BIO scheme (sBIO) that we use in our experiments: a unique I tag is used for each of the output tags; so if the original tag set is $T$, the tags generated by the model are $\overline T \equiv T \cup \{I,O\}$.}
 \vspace{-0.2in}
 \label{tab:sent-example}
 \end{table*}

Data scarcity leads us to consider high-performing encoder-decoder based LLMs. We address deployment concerns by considering distillation of such models into much smaller SL architectures, for instance Bi-Directional Long Short Memory (BiLSTM) \citep{lstm} units, through the use of both labelled and unlabelled data.

A standard distillation approach, knowledge distillation (KD) \citep{https://doi.org/10.48550/arxiv.1503.02531}, requires access to the probability that the teacher network assigns to each of the possible output tags. This probability distribution is typically unavailable at inference time for LLMs; thus, distillation of encoder-decoder models needs to resort to pseudo-labels:\footnote{In this paper, we refer to distillation with pseudo-labels as the process by which a student model is trained on the one-hot labels (and only those labels) generated by a teacher model on an unlabeled dataset. We wish to distinguish this from KD, in which the probability distribution over labels is also used. See also \citet{https://doi.org/10.48550/arxiv.2010.13002}.} the student is trained on the one-hot labels that the teacher assigns to examples in an unlabelled dataset. This prevents the student model from learning those relationships among the probabilities of the incorrect classes that the teacher has learned. Similar arguments apply to decoder-only models.

In this paper, we propose \textbf{SenT$'$}, a simple modification of the \emph{Simplified Inside Sentinel$+$Tag} (\textbf{SenT}) format by \citet{RamanEtAl:22}. We combine our target sequence format with a scoring mechanism for decoding, which we collectively call \textbf{SenTScore}. This combination results in an effective framework that allows us to employ a language model to perform sequence labelling and knowledge distillation. We show that \textbf{SenTScore} is an hallucination-free decoding scheme, and that even with smaller models it outperforms the original \textbf{SenT} format across a variety of standard SL datasets.

Our proposed \textbf{SenTScore} method defines a sequence of scores over the output tags that can be aligned with those generated by the sequence tagging student network, making KD possible. We find an advantage in terms of performance in using KD as opposed to just pseudo-labels as a distillation objective, especially for smaller distillation datasets.

In sum, our contributions are:
\begin{itemize}
    \item A new, hallucination-free, inference algorithm for sequence labelling with encoder-decoder (and possibly decoder only) transformer models, \textbf{SenTScore}, that achieves new state-of-the-art results on multiple English datasets.
    \item Empirical evidence showing an advantage of \textbf{SenTScore} when distilling into a smaller student model. This approach is particularly promising in the few-shot
    setting, which makes it even more appealing and practical.
\end{itemize}

\section{Related work}

Using LLMs to perform sequence tagging is discussed by \citet{athiwaratkun-etal-2020-augmented,yan-etal-2021-unified-generative,https://doi.org/10.48550/arxiv.2101.05779,
https://doi.org/10.48550/arxiv.2110.07298,xue2022byt5} and  \citet{RamanEtAl:22}. While these previous works have minor differences in the prompting format of the models, all but the last one include input tokens as part of the target sequence. Different from our work, all previous models are prone to hallucinate.

Distillation refers to training a small student model from scratch using supervision from a large pretrained model \citep{10.1145/1150402.1150464,https://doi.org/10.48550/arxiv.1503.02531}. Distillation of transformer-based models for different NLP tasks is typically discussed in the context of encoder-only models (e.g. \citealp{ https://doi.org/10.48550/arxiv.1903.12136,mukherjee-hassan-awadallah-2020-xtremedistil,jiao-etal-2020-tinybert}), with a few exceptions looking at distillation of decoder-only models (e.g. \citealp{https://doi.org/10.48550/arxiv.2112.10684}). 

In this paper we will discuss two approaches to distillation: \emph{pseudo-labels} and \emph{knowledge distillation} (KD). In the first approach the student model is trained on the hard labels generated by the teacher on some (unlabelled) dataset. In the second approach additional soft information provided by the teacher is used: typically the probability distribution the teacher assigns to the labels.

In the context of sequence labelling, using pseudo-labels allows us to perform distillation on any teacher-student architecture pair. KD, on the other hand, requires access to the teacher's probability distribution over the output tags. These are not usually available in language models for which the output distribution is over the whole vocabulary of tokens.
We are not aware of other works which modify the decoder inference algorithm to generate such probabilities. However, there is recent work distilling internal representations of the teacher model, with the most closely related work to us being \citet{mukherjee-hassan-awadallah-2020-xtremedistil}. In that work the authors distill a multilingual encoder-only model into a BiLSTM architecture using a two-stage training process. This two-stage process, however, assumes a large unlabelled set for distilling internal model representations, embedding space, and teacher logits, and another significant amount of labelled data for directly training the student model using cross-entropy loss. 

\section{Datasets}
We select seven English datasets that have been used in recent work on slot labelling: ATIS \citep{hemphill-etal-1990-atis}, SNIPS   \citep{https://doi.org/10.48550/arxiv.1805.10190}, MIT corpora (Movie, MovieTrivia, and Restaurant)\footnote{The MIT datasets were downloaded from: \url{https://groups.csail.mit.edu/sls/} }, and the English parts of mTOP \citep{li-etal-2021-mtop} and of mTOD \citep{schuster-etal-2019-cross-lingual}.  Some statistics about the datasets are shown in Table~\ref{tab:datasets}. Some of these datasets (ATIS, SNIPS, mTOP, and mTOD) come from dialogue-related tasks, while the MIT ones have been used for NER.

We use the original training, development, and test sets of the SNIPS, mTOP, and mTOD datasets. For the ATIS dataset we use the splits established in the literature by \citet{atis-split}, in which a part of the original training set is used as the dev set. Similarly, we follow \citet{RamanEtAl:22}\footnote{Private communication with authors} to obtain a dev set out of the original training set for each of the MIT datasets.

We notice that all datasets, with the exception of MovieTrivia, contain some duplicates. Among these, all apart from Restaurant contain examples in the test set that are also duplicated in the train and dev sets. This happens for fewer than 30 instances, with the exception of mTOD, where more than 20\% of the test set examples are also found in the train and dev sets. How these duplicates are handled varies across the literature; we do not remove duplicates from the datasets used in our main results.

However, for mTOD, we also obtained results on a version of the dataset that was deduplicated as follows: If an example is duplicated, we retain it in the highest priority (defined below) split and removed from the others. To ensure the test set is as close as possible to the original test set, we order the splits in ascending order of priority as follows: test, dev, and train. We found that the F1 scores on the deduped mTOD dataset are within 0.5 points those on the original mTOD dataset across all experiments; as such, we do not report the deduped results in the following sections.

In addition to covering different domains, there are noticeable differences across the datasets in terms of the number of tags and the number of labelled examples for evaluation and testing, as can be seen in Table~\ref{tab:datasets}. This set of seven datasets allows us to gather robust empirical evidence for the proposed work that we present in what follows.  

\begin{table}[t]
\centering
\small
\renewcommand{\arraystretch}{1.25}
\begin{tabular}{l|c|c|c|c}
    Datasets & \# tags & \# train & \# dev & \# test \\
    \hline 
    ATIS    & 83      & 4478     & 500    & 893     \\
    SNIPS  & 39      & 13084    & 700    & 700     \\
    MovieTrivia & 12      & 7005     & 811    & 1953    \\
    Movie & 12      & 8722     & 1053   & 2443    \\
    Restaurant & 8       & 6845     & 815    & 1521    \\
    mTOP (en) & 75      & 15667    & 2235   & 4386    \\
    mTOD (en)
    & 16      & 30521    & 4181   & 8621   
    \end{tabular}

\caption{Number of examples per partition and number of unique tags in the SL datasets we used.}
\label{tab:datasets}
\end{table}

\section{Score-based sequence labelling}
Using LLMs for sequence tagging requires reframing the problem as a sequence-to-sequence task.
In \citet{RamanEtAl:22}, the strategy that proved the most effective, at least when applied to the mT5 encoder-decoder architecture, was the \emph{Simplified Inside Sentinel$+$Tag} (\textbf{SenT} in this paper). In this format (see Table \ref{tab:sent-example}), the original text is first tokenized according to some pretokenization strategy (whitespace splitting for all the datasets considered), and each of the tokens is prepended with one of the extra token strings provided by mT5 (the \emph{sentinel} tokens). The resulting concatenation is then tokenized using the mT5 tokenizer and fed to the encoder-decoder model. The output that the decoder is expected to generate is the same input sequence of special token strings, which are now alternated with the tags corresponding to the original tokens.

Given the set $T$ of string labels to be used to annotate a span of text, the scheme used to associate tags across tokens is a modification of the standard BIO scheme: we use $t\in T$ for any token that starts a labelled span, a single tag I for each token that \emph{continues} a labelled span, and O to tag tokens that do not belong to labelled spans.
We refer to this scheme as \emph{Simplified Inside BIO} (sBIO), and we indicate with $\overline T\equiv T \cup \{I, O\}$ the tag set associated to it.

\citet{RamanEtAl:22} argue that the success of SenT can be attributed to two factors: 1) on the one hand, the use of sentinel tokens mimics the denoising objective that is used to pretrain mT5; 2) on the other hand, when compared to other decoding strategies, SenT does not require the decoder to copy parts of the input sentences and also produces shorter outputs. Both these facts supposedly make the task easier to learn and reduce the number of errors from the decoder (\emph{hallucinations}, as  they are often referred to in the literature).

We remark however that any output format among those described in the literature can be made completely free of hallucinations by constraining decoding (either greedy or beam search based) through a finite state machine enforcing the desired output format (see for instance \citealp{https://doi.org/10.48550/arxiv.2010.00904}). In what follows we describe our proposed decoding approach that builds on this previous work.
\subsection{SenTScore}
Regardless of possible constraints imposed during generation, both \textbf{SenT} and the other algorithms described in \citet{RamanEtAl:22} use the decoder autoregressively at inference time to generate the target sequence. Since generation proceeds token by token and the textual representation of a tag is a variable length sequence of tokens, it is nontrivial to extract the scores and probabilities that the model assigns to individual tags. 

We propose a different approach to inference, one in which the decoder is used to score sequences of tags. For this purpose, we consider a sequence tagging task with a label set $T$, and the associated sBIO tag set $\overline T$. Given an input sentence $S$, we use a pre-tokenizer (such as whitespace splitting) to turn $S$ into a sequence of token strings $x_1\ldots x_L$, of size $L$. The SenT format is obtained by interleaving these tokens with special token strings to obtain the input string $S_{\textrm{in}}=s_0x_1s_1\ldots x_L$. We use juxtaposition to indicate string concatenation. In what follows, we will work with \textbf{SenT$'$}, a modification of SenT in which an additional special token is appended at the end, $S_{\textrm{in}}\leftarrow S_{\textrm{in}}s_L$. The reason for doing this will become clear in what follows.

The valid output strings that can be generated by the decoder are the $|\overline{T}|^L$ sequences of the form $S_{\textrm{out}}=s_0 t_1s_1\ldots t_Ls_L \in \mathcal O$ where $t \in \overline T\equiv T\cup\{I, O\}$ consistent with the sBIO scheme convention. The encoder-decoder model can be used to calculate the log-likelihood of each of such strings ${\log\mathcal L_\theta}(S_{\textrm{out}};S_{\textrm{in}})$, where $\theta$ represents the model parameters, and the best output will be:
\begin{equation*}
S_{\textrm{out}}^* =\arg\max_{S\in \mathcal O} {\log\mathcal L}(S;S_{\textrm{in}})
\end{equation*}
Exact inference is infeasible but can be approximated using beam search as described in Algorithm~\ref{fig:sentscorer-algo}. The outputs of the algorithm are the top-K output strings and the score distribution associated with each of the output tags. As is evident from Table~\ref{tab:sent-example}, it is simple to map back the final output string $S^*$ to the sequence of output tags and labelled spans.

\begin{algorithm}[t]
  \small
  \begin{algorithmic}
    \Require{Encoder-decoder parameters $\theta$, input $S_{\textrm{in}}$ with $L$ tokens, sBIO tag set $\overline T$, beam size $K$}
    \Ensure{Approximate top-$K$ output sequences $\mathcal B_{\textrm{text}}$ and their sBIO tag scores, $\mathcal B_{\textrm{scores}}$}
    \State{$\mathcal B_{\textrm{text}} \gets [\,s_0\,]_{i=1\ldots K}$}
    \State{$\mathcal B_{\textrm{scores}} \gets [\,[~]\,]_{i=1\ldots K}$}
    \For{$i = 1$ to $L$}

    \State{$\mathcal H \gets [\,z\,t\,s_i\,]_{z \in \mathcal B_{\textrm{text}},\, t \in \overline T}$} \Comment{Generate hypotheses}
    \State{$\mathcal S \gets [\,{\log\mathcal L}_\theta(h; S_{\textrm{in}})\,]_{h \in \mathcal H}$} \Comment{Score hypotheses}

    \State{$\Pi \gets \textrm{K-argsort}\, \mathcal S$} \Comment{top-K args}
    \State{$\mathcal B_{\textrm{text}} \gets \textsc{Take}(\mathcal H; \Pi)$} \Comment{Update text beam}
    \State{$\widetilde{\mathcal S} \gets {\textsc{Reshape}}(\mathcal S; K, |\overline T|);$} \Comment{Reshape scores}
    \For{$k = 0$ to $K-1$} \Comment{Update score beam}
    \State{$\widetilde k \gets \Pi[k] \mod K$}
    \State{$\mathcal B_{\textrm{scores}}[k] \gets \textsc{Append}(\mathcal B_{\textrm{scores}}[\widetilde k];\mathcal S[\widetilde k])$}
    \EndFor
    \EndFor
    \State{\Return{ $\mathcal B_{\textrm{text}}$, $\mathcal B_{\textrm{scores}}$}}
  \end{algorithmic}
  \caption{\label{fig:sentscorer-algo} SenTScore beam search }
\end{algorithm}

At decoding time the output string is initialized with the first sentinel token $s_0$.
At the $i$-th step, \textbf{SenTScore} uses the model likelihood to score each of the $|\overline T|$ possible continuations of the output sequence
\begin{equation}\label{eq:next}
t\,s_i~~{\textrm{with}}~~t\in \overline{T},
\end{equation}
picks the highest scoring one, and keeps track of the score distribution. $s_i$ in Eq.~\ref{eq:next}, the \emph{next} sentinel token, plays the crucial role of an EOS token at each step. This is needed to normalize the probability distribution: the likelihood of the string $s_0t_1\ldots s_{k-1}t'_k$ is always bounded by that of the string $s_0t_1\ldots s_{k-1}t_k$ if $t$ is a prefix of $t'$, and we would never predict $t'$ as a continuation of $s_0t_1\ldots s_{k-1}$. This explains why we prefer using \textbf{SenT$'$} over \textbf{SenT}.

Finally, while \textbf{SenTScore} changes the inference algorithm, the finetuning objective we use throughout is still the original language modelling one.

\subsection{Distillation}\label{subsec:distillation}
The main advantage of \sentscore is in the distillation setting. 
At each inference step, the algorithm assigns a likelihood to each sBIO tag. This distribution can be used to train the student network by aligning it to the teacher's pre-softmax logits, in a standard knowledge distillation setup.

In detail, given an input sequence $S_{\textrm{in}}$, let $(\mathbf{y}^*_i)_{i=1\ldots L}$ be the sequence of sBIO output tags (as $|\overline T|$-dimensional one-hot vectors) as inferred by the teacher model, and let $(\mathbf{u}^*_i)_{i=1\ldots L}$ (also $|\overline T|$-dim. vectors) be the associated sequence of log-likelihoods. We indicate with $\mathbf{p}^*_i$ the probability obtained by softmaxing $\mathbf{u}^*_i$ and by $\mathbf{q}_i$ the output of the softmax layer from the student. The contribution of each of the tags to the distillation objective that we use to train the student sequence tagger is
\begin{equation}\label{eq:distloss}
-\sum_k(y^*_i)_k \log \mathbf (p^*_i)_k + \lambda_{KL}\, KL(\mathbf p^*_i|| \mathbf q_i)\,.
\end{equation}
The first term is the standard cross-entropy contribution from the pseudo-labels, while the second is the knowledge distillation term, implemented with a KL divergence with $\lambda_{KL}$ its associated positive weight. 

We stress that we are allowed to write the second term only because \textbf{SenTScore} provides us with the tag scores. This is not the case for any of the formats proposed in \citet{RamanEtAl:22} or, as far as we know, elsewhere.\footnote{Strictly speaking the student defines $p(\cdot|t_{1}^*\ldots t_{i-1}^*;S_{\textrm{in}})$ (star means predicted) while $\mathbf q_i$ corresponds to $p(s_0 t_{1}^*\ldots t^*_
{i-1} s_
{i-1}\cdot|S_{\textrm{in}})$. This discrepancy is resolved by the invariance of the softmax under constant shifts of its arguments.}

\section{Experimental settings}
We evaluate the models by computing the \textbf{F1} score on the test set of each dataset. \textbf{F1} is calculated following the CoNLL convention as defined by \citet{tjong-kim-sang-de-meulder-2003-introduction}, where an entity is considered correct iff the entity is predicted exactly as it appears in the gold data. We show micro-averaged \textbf{F1} scores.

The first set of experiments we performed are intended to investigate whether our proposed \textbf{SentTScore} approach is competitive with respect to recent results on the same datasets (Table~\ref{tab:sentinel-score}). Our \textbf{SentTScore} model is a pretrained T5-base model (220M parameters) finetuned on each of the datasets.\footnote{All our results are in the greedy setting. We find very small differences in performance by using beam search, while inference time grows considerably.} We trained each model for 20 epochs, with patience 5, learning rate of $10^{-3}$, and batch size 32. We also want to know how the proposed framework compares against the following strong baselines:\\
\textbf{BiLSTM:} Our first baseline is a BiLSTM tagger \cite{https://doi.org/10.48550/arxiv.1603.01360}.\footnote{We do not include a CRF layer.} 
The BiLSTM has a hidden dimension of size $200$. Its input is the concatenation of 100d pretrained GloVE6B embeddings \citep{pennington-etal-2014-glove} from \citet{glove6b} with the 50d hidden state of a custom character BiLSTM. We trained each model for 100 epochs, with patience 25, learning rate of $10^{-3}$, and batch size 16.\\
\textbf{BERT:} We finetune a pretrained BERT-base cased model \citep{devlin-etal-2019-bert} (110M parameters) for the SL task and report results for each of the seven datasets. While we consider BERT a baseline model, we note that this pretrained architecture continues to show good performance across a wide range of NLP tasks, and for models in this size range BERT is still a reasonable choice. In preliminary experiments we compared results from the case and uncased versions of BERT and we found negligible differences. We decided to use the cased version for all experiments reported here.  We trained each model for 30 epochs, with patience 10, learning rate of $5\times 10^{-5}$, and batch size 64.\\
\textbf{SentT$'$}: The pretrained model is the same as that used for \textbf{SentTScore}.  The goal of this baseline is to assess improvements attributed to our proposed decoding mechanism. This model is also the closest model to prior SOTA. 
The main difference between our results and those in \citet{RamanEtAl:22} is the pretrained model. They used a multilingual T5 model \cite{xue-etal-2021-mt5} with 580M parameters, whereas we use a smaller monolingual version \citep{JMLR:v21:20-074}. 

All the above models were trained with the AdamW optimizer \citep{Loshchilov2017FixingWD}. The best checkpoint for each training job was selected based on highest micro-F1 score on the validation set. All pretrained transformer models are downloaded from \citet{huggingface}.

\begin{table*}[ht]
\centering
\small
\renewcommand{\arraystretch}{1.25}
\begin{tabular}{l|cc|cc|cc|cc||cc}
    \multirow{2}{*}{Dataset} & \multicolumn{2}{c|}{BiLSTM [1M]} & \multicolumn{2}{c|}{BERT [110M]} & \multicolumn{2}{c|}{T5 [220M] (SenT$'$)} & \multicolumn{2}{c||}{T5 (SenTScore)} & \multicolumn{2}{c}{mT5 [580M] (SenT) }\\
          & Perfect    & F1    & Perfect    & F1       & Perfect    & F1     & Perfect    & F1 & Perfect    & F1   \\ \hline 
    ATIS & 89.06 &	95.56 &	88.57 &	95.27 &	86.56 &	94.77 &	\textbf{89.81} &	\textbf{95.99} & \underline{90.07} & 95.96 \\   
    SNIPS & 87.24 & 95.02 & 89.71 & 95.47 & 89.86 & 95.43 & \textbf{91.00} & \textbf{96.07} & 89.81 & 95.53\\
    MovieTrivia & 32.41 & 69.81 & 36.2 & 69.15 & 36.35 & 70.76 & \textbf{39.58} & \textbf{71.99} & \underline{39.85} &	\underline{73.01}\\
    Movie & 69.79 &	86.72 &	69.46 &	85.83 &	71.88 &	87.53 &	\textbf{74.29} & \textbf{88.35} & 72.74 & 87.56\\
    Restaurant & 58.32 & 77.39 & 58.97 & 77.69 & 58.65 & 78.77 & \textbf{63.77} & \textbf{80.91} & 62.93	& 80.39\\
    mTOP (en) & 81.10 & 88.94 & 84.4 & 90.98 & 84.18 & 90.64 & \textbf{86.66} & \textbf{92.29} & 86.56 &	92.28\\
    mTOD (en) & 91.70 & 95.62 & 92.35 & 95.83 & 92.24 & 96.04 & \textbf{92.94} & \textbf{96.24} & \underline{93.19} &	\underline{96.42}\\
\end{tabular}

\caption{Our results comparing BERT-base and a BiLSTM against a T5-base model using SenT$'$ and \textbf{SenTScore} on different SL datasets are shown in the first 4 columns. Number in square brackets are model sizes in terms of number of parameters. Results from \citet{RamanEtAl:22} are copied in the last column. Bold scores represent our best results; underlined scores in the last column highlight those cases in which \citet{RamanEtAl:22} outperforms us.}
\label{tab:sentinel-score}
\end{table*}

\begin{table*}[ht]
\centering
\small

\renewcommand{\arraystretch}{1.25}
\begin{subtable}{.47\linewidth}

\begin{tabular}{l|c|c|c|c}
    
   Dataset - F1  & BiLSTM & BERT & 
    T5 &
    \vtop{\hbox{\strut BiLSTM}\hbox{\strut (distilled)}}\\ \hline 
    ATIS & 79.93 & 79.43 & 85.01 & \textbf{86.75} \\   
    SNIPS & 51.63 & 52.16 & 54.33 & \textbf{57.18}\\
    MovieTrivia & 48.26 & 50.26 & 55.74 & \textbf{57.85}\\
    Movie & 60.82 & 61.80 & 67.09 & \textbf{70.51}\\
    Restaurant & 47.26 & 53.17 & 56.87 & \textbf{61.13}\\
    mTOP (en) & 43.12 & 46.08 & 51.94 & \textbf{54.77}\\
    mTOD (en) & 68.68 & 76.95 & 79.43 & \textbf{82.26}\\
\end{tabular}
\caption{\label{tab:dist100}Gold train/dev split of size 100/50}
\end{subtable}
\quad
\quad
\begin{subtable}{.47\linewidth}

\begin{tabular}{l|c|c|c|c}
    Dataset - F1 & BiLSTM & BERT & 
    T5 &
    \vtop{\hbox{\strut BiLSTM}\hbox{\strut (distilled)}}\\ \hline 
    ATIS & 86.43 & 84.95 & 89.33 & \textbf{90.25} \\   
    SNIPS & 69.19 & 72.77 & 76.34 & \textbf{79.84}\\
    MovieTrivia & 57.64 & 58.41 & 63.60 & \textbf{65.34}\\
    Movie & 73.54 & 73.76 & 77.39 & \textbf{79.20}\\
    Restaurant & 61.62 & 62.97 & 68.52 & \textbf{68.62}\\
    mTOP (en) & 57.22 & 63.28 & 67.73 & \textbf{69.62}\\
    mTOD (en) & 83.46 & 85.51 & 88.68 & \textbf{89.82}\\
\end{tabular}
\caption{\label{tab:dist300}Gold train/dev split of size 300/150}
\end{subtable}

\caption{Distillation results and comparisons with baselines. The distillation results use the full objective function in Eq.~\ref{eq:distloss} with $\lambda_{KL}=1$.}
\label{tab:distill-full}
\end{table*}

\subsection{Distillation experiment}

We apply \textbf{SentTScore} and the loss function described in Section ~\ref{subsec:distillation}, to distill a finetuned T5 model into a BiLSTM architecture to perform sequence tagging. To mimic a low-resource setting, we randomly downsample the train/dev splits of all the datasets. We consider two sets of sizes for these gold train/dev splits: a 100/50 split and a 300/150 one. In both settings the remainder of the original training set is used for the distillation component using pseudo-labels.

We then finetune T5 using the SenT$'$ format on each of these two gold splits. The resulting model is used as the teacher in a distillation setting in which the student is a BiLSTM. The BiLSTM student is trained on the full training set by using the downsampled gold labels, but pseudo-labels and scores generated by the T5 teacher using \textbf{SenTScore} with $K=1$ in the rest of the training data. We use a temperature parameter $\tau$ to rescale the distribution \textbf{SenTScore} defines over $\overline T$. We use $\tau=10$ in all the distillation experiments.

The training schedule we follow is the same we use to train the BiLSTM baseline model, with the only exception that the best checkpoint is selected on the reduced dev set.

\section{Results}\label{sec:results}
 

The comparisons between baselines, SenT$'$, and \textbf{SenTScore} are shown in Table~\ref{tab:sentinel-score}. \textbf{SenTScore} is used with a $K=1$ beam size. Larger beams result in very similar performance and a considerable slowdown of inference time.
\textbf{SenTScore} consistently outperforms SenT$'$ with constrained decoding, and all other baselines. Our intuition is that one advantage of \textbf{SenTScore} comes from the fact that decoding happens tag-wise as opposed to token-wise (as in pure beam search). The last column of Table~\ref{tab:sentinel-score} shows the performance of the SenT implementation of \citet{RamanEtAl:22}. \textbf{Perfect} scores are also reported for completeness. They are evaluated at the sentence level and correspond to the fraction of perfectly predicted examples. However these results are not directly comparable: \citet{RamanEtAl:22} use a different and larger model (mT5-base with 580M parameters) and different optimization details. Nevertheless \textbf{SenTScore} achieves better performance in a majority of cases.

\subsection{Distillation results}

\begin{table*}[ht]
\centering
\small
\renewcommand{\arraystretch}{1.25}
\begin{tabular}{l|cc|cc|cc}
    \multirow{2}{*}{Dataset - F1} & \multicolumn{2}{c|}{No silver} & \multicolumn{2}{c|}{250 silver} 
    & \multicolumn{2}{c}{500 silver}\\
         &  $\lambda_{KL}=0$   & $\lambda_{KL}=1$    &  $\lambda_{KL}=0$  & $\lambda_{KL}=1$  & $\lambda_{KL}=0$ & $\lambda_{KL}=1$    \\ \hline 
    ATIS & 79.93 (0.85) & \textbf{82.35} (0.44) & 83.09 (1.49) & \textbf{84.42} (1.42) & 83.75 (1.74) & \textbf{85.10} (1.54)\\   

    SNIPS & 51.63 (1.25) & \textbf{55.65} (1.38) & 54.34 (0.71) & \textbf{56.02} (1.71) & 55.66 (1.21) & \textbf{57.00} (1.17)\\
    
    MovieTrivia & 48.26 (0.95) & \textbf{51.86} (1.09) & 53.11 (1.26) & \textbf{55.19} (0.50) & 53.97 (1.55) & \textbf{56.00} (0.38)\\
    
    Movie & 60.82 (0.67) & \textbf{64.20} (1.07) & 67.04 (0.66) & \textbf{69.41} (0.66) & 67.73 (1.28) & \textbf{70.12} (0.60)\\
    
    Restaurant & 47.26 (0.83) & \textbf{50.19} (0.81) & 54.24 (1.17) & \textbf{56.20} (0.94) & 56.29 (1.11) & \textbf{57.95} (0.88)\\
    
    mTOP (en) & 43.12 (1.84) & \textbf{46.43} (1.01) & 46.68 (2.31) & \textbf{49.08} (1.65) & 49.57 (0.47) & \textbf{50.33} (2.17)\\
    
    mTOD (en) & 68.68 (2.68) & \textbf{70.40} (0.97) & 76.12 (1.07) & \textbf{77.86} (1.45) & 77.86 (0.85) & \textbf{79.77} (0.82)\\
\end{tabular}
\caption{\label{tab:abl100} Distillation experiments with varying silver dataset size and ablation of the KD term in Eq.~\ref{eq:distloss}. The gold data split is the same as in Table~\ref{tab:dist100}, with train/dev sizes of 100/50. The numbers in parentheses represent the standard deviation of the scores obtained by varying all the random seeds that appear at training time: BiLSTM weight initialization, batch scheduling, and the choice of the silver data set.}
\end{table*}

\begin{table*}[ht]
\centering
\small
\renewcommand{\arraystretch}{1.25}
\begin{tabular}{l|cc|cc|cc}
    \multirow{2}{*}{Dataset - F1} & \multicolumn{2}{c|}{No silver} & \multicolumn{2}{c|}{250 silver} 
    & \multicolumn{2}{c}{500 silver}\\
           &  $\lambda_{KL}=0$   & $\lambda_{KL}=1$    &  $\lambda_{KL}=0$  & $\lambda_{KL}=1$  & $\lambda_{KL}=0$ & $\lambda_{KL}=1$     \\ \hline 
    ATIS & 86.43 (1.09) & \textbf{88.42} (0.48) & 89.15 (0.65) & \textbf{89.39} (0.49) & 89.73 (0.66) & \textbf{89.98} (0.31)\\   

    SNIPS & 69.19 (0.74) & \textbf{73.06} (0.54) & 72.11 (1.11) & \textbf{75.02} (1.16) & 73.99 (0.81) & \textbf{75.73} (1.47)\\
    
    MovieTrivia & 57.64 (0.45) & \textbf{60.30} (0.34) & 60.25 (0.37) & \textbf{62.11} (0.54) & 61.38 (0.46) & \textbf{62.89} (0.52)\\
    
    Movie & 73.54 (0.40) & \textbf{76.30} (0.33) & 75.88 (0.44) & \textbf{76.96} (0.44) & 76.52 (0.58) & \textbf{77.58} (0.26)\\
    
    Restaurant & 61.62 (0.43) & \textbf{63.78} (0.27) & 64.33 (0.90) & \textbf{65.33} (0.66) & 65.20 (0.80) & \textbf{66.24} (0.63)\\
    
    mTOP (en) & 57.22 (0.73) & \textbf{60.36} (0.50) & 60.81 (0.92) & \textbf{62.70} (0.69) & 62.02 (0.94) & \textbf{64.37} (0.84)\\
    
    mTOD (en) & 83.46 (0.59) & \textbf{85.52} (0.20) & 86.35 (0.40) & \textbf{87.08} (0.50) & 86.82 (0.33) & \textbf{87.89} (0.40)\\
\end{tabular}
\caption{\label{tab:abl300} Distillation experiments with varying silver dataset size and ablation of the KD term in Eq.~\ref{eq:distloss}. The gold data split is the same as in Table~\ref{tab:dist300}, with a train/dev size given by 300/150. All experimental details are common with Table~\ref{tab:abl100}.}
\end{table*}

Tables~\ref{tab:dist100} and \ref{tab:dist300} show the result of the distillation experiments with 100/50 and 300/150 train/dev gold splits, respectively. While a BiLSTM tagger trained on the gold data significantly underperforms a finetuned T5-base model, once the BiLSTM is distilled on the silver data generated using SenTScore, it outperforms even the original teacher model. We notice that the difference between student and teacher decreases for larger gold set size, suggesting that the effect is related to regularization properties of the distillation process. A similar phenomenon has been observed elsewhere, for instance in \citet{bornagain} albeit with teacher and student sharing the same architecture. 

In order to isolate the benefits of training the teacher model using KD as opposed to just pseudo-labels, we perform a set of ablation studies. For each dataset, we distill a BiLSTM student on a training set $\mathcal T = \mathcal G \cup \mathcal S$, where $\mathcal G$ is the original gold set and $\mathcal S$ is a random sample from the complement of $\mathcal G$. We choose $|\mathcal S| = 0, 250, 500$. 
The student is distilled using Eq.~\ref{eq:distloss} with two choices of the loss multipliers: $\lambda_{KL}=1$ and $\lambda_{KL}=0$. The first setting is the same used in Tables~\ref{tab:dist100} and \ref{tab:dist300}, while the second drops the KD loss and only keeps the pseudo-labels for distillation. Whenever pseudo-labels and scores are used, they are generated by the SenTScore algorithm.

\begin{figure*}[h!]
\centering
\includegraphics[scale=0.66]{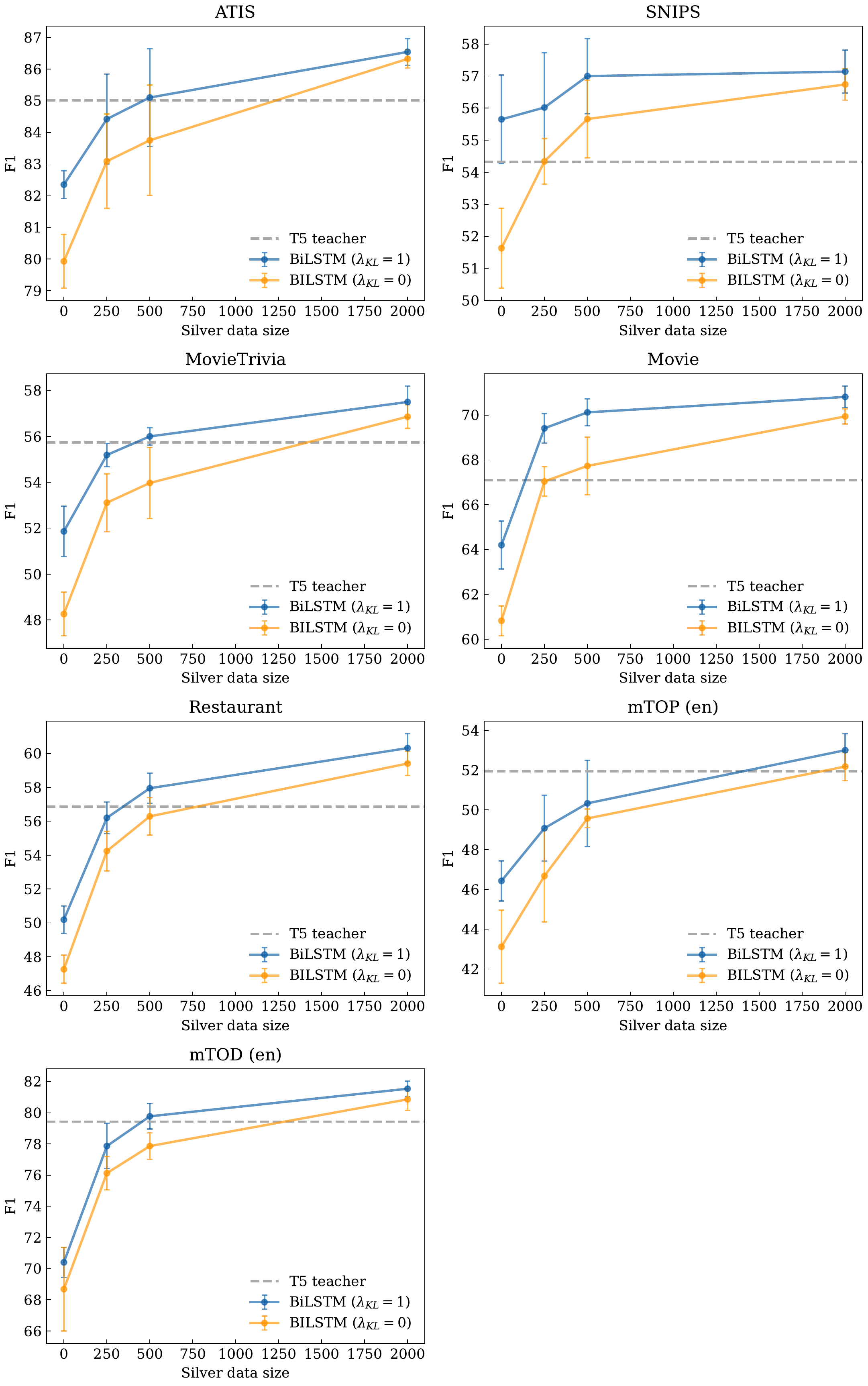}
\caption{A graphical representation of the distillation results in Table~\ref{tab:dist100} (100/50 gold train/dev split) as a function of the size of the silver dataset. Knowledge distillation using SenTScore generated scores outperforms pseudo-labels.}\label{fig:all}
\end{figure*}

The results are shown in Tables~\ref{tab:abl100} and \ref{tab:abl300}. We see a consistent trend in which KD outperforms training the student using only pseudo-labels. This in particular motivates SenTScore as an inference algorithm. The results also show that for our choice of teacher and student architectures, and datasets, the gap between KD and pseudo-labels is reduced when more silver data are used. Figure~\ref{fig:all} further explores the relationship between amount of pseudo-labeled data and gains from KD with $|\mathcal S| = 0, 250, 500, 2000$.
The trend with more pseudo-labeled data remains unchanged. 
\section{Limitations and future work}
A reasonable critique to our focus on real-world constraints is the simple fact the datasets we are using are not real-world ones. From noise to tokenization choices, many issues arise when considering datasets outside of the academic domain. However, we believe our methods are simple enough to be applicable to real-world scenarios and our results to be independent of these various subtleties.

Some issues that could be addressed in future work have to do with the exploration of even larger models and different architectures such as decoder-only ones \cite{RadfordEtAl:18, RadfordEtAl:19, brown2020gpt, ZhangEtAl:22, chowdhery2022palm, gpt-neo}. We note, however, that in all our experiments we finetune all the weights of the pretrained models we use. When using extremely large models this becomes impractical. Recent work \cite{minibert} suggests that KD with compact encoder-only student models, such as BERT, is a promising avenue for further research. Exploring the pure few-shot scenario, or only finetuning a subnetwork, for instance by using adapters \`{a} la \citealp{adapters}, would be also interesting.

\section{Conclusion}
Real-time systems need to find a trade-off between performances and computing resources, the latter constraint coming either from budget or some other service requirement. Such trade-offs become particularly evident with large pretrained transformer models, which achieve SOTA results on many NLP tasks at the cost of being extremely hard and expensive to deploy in a real-world setting.

The standard solution for this is distillation. In this paper we have revisited these issues for the SL task, which is often the first crucial step in many real-world NLP pipelines. We propose a new inference algorithm, SenTScore, that allows us to leverage the performance of arbitrarily large encoder-decoder transformer architectures by distilling them into simpler sequence taggers using KD as opposed to just pseudo-labelling.

\section*{Ethical considerations}
The intended use of our proposed approach is related to sequence labelling tasks where there are latency constraints and limited labelled data available. While it is not impossible to identify potential misuses of this technology, it is not immediately clear what those malicious uses would be. On the contrary, this paper contributes to the body of work investigating efficient solutions for deployment of live systems.

\subsection*{Computing infrastructure and computational budget}
All of our experiments were run on single V100 GPU machines with 32GB. The most expensive experiments relate to finetuning a model, including best checkpoint selection. In this case, the running time is directly related to the dataset size. For the experiments using the full train/dev set, running time varies from 45 minutes (mATIS corpus) to a few hours (mTOD corpus) for a T5-base model. 
Training a model takes, on average,  around 4 iterations per second with batch size 32. For the generation of pseudo-labels, we did not implement batch processing and it takes around 0.15 seconds to annotate each sample.
\nocite{sls-corpora}
\bibliography{anthology,custom}




\end{document}